\documentclass[preprint,12pt]{elsarticle}

\usepackage{amssymb}
\usepackage{mathtools}
\usepackage{amsthm}
\usepackage[noend]{algorithmic}
\usepackage{algorithm}
\usepackage{tabularx}
\usepackage[table]{xcolor}
\theoremstyle{plain}

\newtheorem{proposition}{Proposition}
\newtheorem{lemma}{Lemma}

\theoremstyle{definition}
\newtheorem{definition}{Definition}

\theoremstyle{remark}

\journal{}

\begin{document}

\begin{frontmatter}



\title{Domain Gap Estimation for Source Free Unsupervised Domain Adaptation with Many Classifiers}


\author[inst1]{Ziyang Zong}

\affiliation[inst1]{organization={School of Artificial Intelligence},
            Organization={Nanjing University of Information Science and Technology},
            addressline={}, 
            city={Nanjing},
            postcode={210044}, 
            state={Jiangsu},
            country={China}}

\author[inst2]{Jun He \corref{cor1}} 
\ead[url]{https://sites.google.com/site/hejunzz/}
 \cortext[cor1]{Corresponding author.}

\author[inst3]{Lei Zhang}
\author[inst1]{Hai Huan}

\affiliation[inst2]{organization={Independent Researcher},
            city={Cupertino},
            postcode={95014}, 
            state={CA},
            country={USA}}
            
\affiliation[inst3]{organization={School of Electrical and Automation Engineering},
            Organization={Nanjing Normal University},
            addressline={}, 
            city={Nanjing},
            postcode={210023}, 
            state={Jiangsu},
            country={China}}

\begin{abstract}
 In theory, the success of unsupervised domain adaptation (UDA) largely relies on domain gap estimation. However, for source free UDA, the source domain data can not be accessed during adaptation, which poses great challenge of measuring the domain gap. In this paper, we propose to use many classifiers to learn the source domain decision boundaries, which provides a tighter upper bound of the domain gap, even if both of the domain data can not be simultaneously accessed. The source model is trained to push away each pair of classifiers whilst ensuring the correctness of the decision boundaries. In this sense, our many classifiers model separates the source different categories as far as possible which induces the maximum disagreement of many classifiers in the target domain, thus the transferable source domain knowledge is maximized. For adaptation, the source model is adapted to maximize the agreement among pairs of the classifiers. Thus the target features are pushed away from the decision boundaries. Experiments on several datasets of UDA show that our approach achieves state of the art performance among source free UDA approaches and can even compete to source available UDA methods.
\end{abstract}




\end{frontmatter}


\section{Introduction}
Unsupervised domain adaptation (UDA) aims for adapting the source domain knowledge to the target domain~\cite{ganin2015unsupervised,long2016unsupervised}. UDA brings a new learning paradigm for the challenging task where the real world samples are of huge volume but hard to annotate.
For example, in the application of semantic segmentation in self-driving, even though real world video frames taken by camera are produced in unprecedented speed, people can train a source model by using a car game simulator which can produce the well annotated frames, and then adapt the source model to the real world video frames by using UDA approaches to mitigate the domain gap.
Obviously, a large domain gap inevitably exists between the source and the target domains since the samples of each domain are collected in different ways.

In theory, the success of UDA largely relies on domain gap estimation~\cite{ben2010theory}, saying that the more accurate and robust the domain gap is estimated, the lower target error could be achieved. 
Previous work has addressed this problem by maximizing the classification discrepancy between bi-classifier by leveraging both domain data simultaneously~\cite{saito2018maximum}. However, this setup breaks up in the source free setting. Source free UDA is popular in real world scenarios due to data privacy~\cite{liang2020we}. In this setting, measuring the domain gap becomes more challenging. 
On the other hand, the recent proposed source free UDA approaches are trying to bring the source domain classification information into the target domain by fixing the trained classifier, such as SHOT~\cite{liang2020we}. 
Specially, besides the conditional entropy loss and the marginal entropy loss which are always adopted in unsupervised learning literature~\cite{krause2010discriminative, DBLP:journals/corr/Springenberg15}, these approaches usually depend on the pseudo labels obtained by clustering the target features~\cite{liang2020we, yang2021generalized}, which are hard to achieve higher accuracy. It is because clustering the target features can only rely on the learned source domain feature representation which may be far away from the true representation of target domain. 


\begin{figure*}[!t]
\vskip 0.2in
\begin{center}
\centerline{\includegraphics[width=0.95\textwidth]{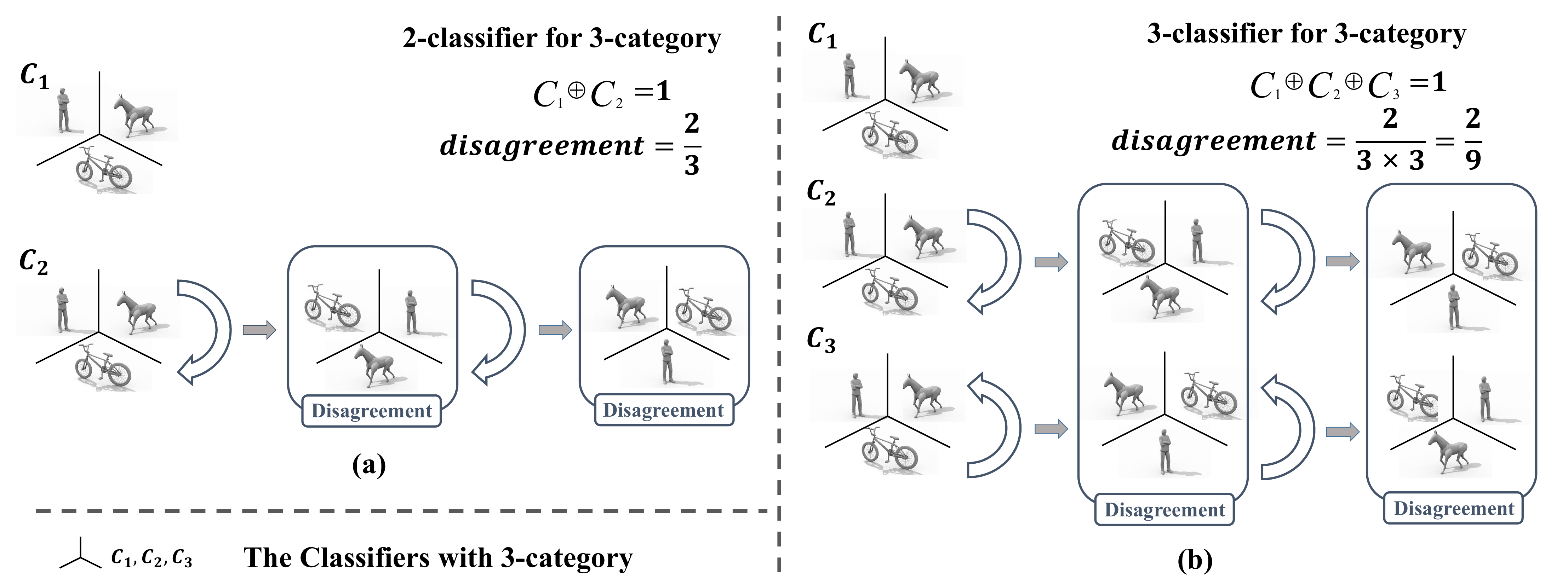}}
\caption{Comparison of disagreement ratio between bi-classifier model and 3-classifier model for 3-category task. 3-classifier model gives a lower disagreement ratio than bi-classifier model, which induces a tighter domain gap estimation.}
\label{fig:introduction}
\end{center}
\vskip -0.2in
\end{figure*}

The seminal theoretical work~\cite{ben2010theory} proposed to use the symmetric difference hypothesis space $\mathcal{H}\Delta\mathcal{H}$ to measure the domain gap as Equation~\eqref{eq:hh_divergence} which mainly considers the disagreement between two hypotheses in both domains. 
\begin{small}
  \begin{equation}
  \label{eq:hh_divergence}
    \begin{split}
      d_{\mathcal{H}\Delta\mathcal{H}}(\mathcal{D}_S,\mathcal{D}_T)=2\sup_{h, h'\in \mathcal{H}\Delta\mathcal{H}}\left |P_{x\sim\mathcal{D}_S}[h(x)\neq h'(x)]-P_{x\sim\mathcal{D}_T}[h(x)\neq h'(x)]\right |
    \end{split}
  \end{equation}
\end{small}
Then theoretically, a good algorithm of UDA should bound the gap tightly so that the following adaptation could optimize the model to minimize the domain gap correctly. When the source data are available in adaptation, the bi-classifier model, such as MCD~\cite{saito2018maximum} and BCDM~\cite{li2020bi}, can actually search the whole $\mathcal{H}\Delta\mathcal{H}$ space to determine the upper bound of the domain gap. However this bound of bi-classifier model will become much looser when source data are unavailable in adaptation for practical UDA tasks, because the classifiers of the bi-classifier model are always fixed to maintain the source domain knowledge~\cite{liang2020we}.
Figure~\ref{fig:introduction}~(a) demonstrates that for 3-category task, the disagreement ratio of bi-classifier model is $\frac{2}{3}$, saying that there are two disagreements satisfying $C_1\oplus C_2=1 $ among the three configurations. This ratio is larger than $\frac{1}{2}$, the ratio of 2-category task. If the classification task has more categories, the disagreement ratio closes to 1. In this sense, the bi-classifier model provides a much looser upper bound in source free setting. 


Can the domain gap be estimated more accurate and more stable even in the scenario of source free? The answer is Yes.
Figure~\ref{fig:introduction}~(b) demonstrates that for the same 3-category task,
if we use a 3-classifier model in source free setting, there are two disagreements satisfying $C_1\oplus C_2 \oplus C_3=1 $ among the 9 configurations which induces a much lower disagreement ratio $\frac{2}{9}$.  Obviously, 3-classifier model provides a much tighter upper bound of domain gap than bi-classifier. Moreover, in domain adaptation, 3-classifier model requires the agreements among all 3 classifiers which is more rigorous than bi-classifier model.
This motivates us to explore how many classifiers should a model be best for bounding the domain gap in source free UDA. 
In this paper, we develop a theory of why many classifiers can bound the domain gap much tighter in source free setting and propose a framework of source free UDA via many classifiers (DAMC). 
In our model, the tighter upper bound is guaranteed on the assumption that the decision boundaries of different classifiers should be placed as far as possible in the source domain whilst ensuring the correctness, which induces the largest possibility of disagreement of many classifiers in the target domain. Thus in this sense, the transferable source domain knowledge is maximized. Our DAMC framework is composed of source model pre-training and target domain adaptation. For pre-training the source model, the decision boundaries of each pair of classifiers are trained to be pushed away whilst ensuring the correctness. For target domain adaptation, the source model is adapted to maximize the agreement among pairs of the classifiers. Thus the target features are pushed away from the decision boundaries.
%

The contributions of this paper are threefold.
\begin{itemize}
    \item We propose to use many classifiers as a strong domain discriminator for source free UDA. Importantly, from the perspective of metric learning, we answer why using many classifiers can measure the domain gap more accurately and stably than bi-classifier in the source free setting.
    \item We not only provide the optimal number of classifiers for a specific source free UDA task, but also discuss the trade-off between domain gap estimation and the optimal number of classifiers when the task has many categories. 
    \item A framework of source free unsupervised domain adaptation via many classifiers  (DAMC) is proposed. We introduce a novel and effective pretraining method for the source domain which pushes away the decision boundaries of each pair of classifiers whilst ensuring the correctness. For target domain adaptation, we propose a novel loss function to measure the agreements among many classifiers.
\end{itemize}

The remaining sections are organized as follows. Section~\ref{sec:rel_work} presents the related work on domain gap estimation, unsupervised domain adaptation, and source free UDA. In Section~\ref{sec:dmac_theory}, we extend the theory of domain gap estimation to the source free scenario and provide the reason of why should use many classifiers for source free UDA. In Section~\ref{sec:algorithm}, the framework of DAMC is presented, which is composed of the source model pre-training 
module and the target adaptation module. Section~\ref{sec:experiments} validates our theory and the proposed source free UDA framework DAMC by examining its performance on popular UDA benchmark datasets. Section~\ref{sec:conclusion} concludes our work.


\section{Related Work}
\label{sec:rel_work}
We pay close attention to unsupervised domain adaptation(UDA).
In order to bound the transferal error between different domains, symmetric difference hypothesis space $\mathcal{H}\Delta \mathcal{H}$~\cite{ben2010theory} which exploits XOR to measure the disagreement between two hypothesises was introduced to define the upper bound of the target error for the 2-category problem, which laid a theoretical foundation for subsequent work. For the multi-category DA problem, 
theoretical works~\cite{zhang2020unsupervised, lee2021compact} were further expanded on top of~\cite{ben2010theory}, which exploit the expectation of XOR for estimating the inconsistent of the multi-category problem.

A successful family of research on UDA adopts the adversarial training framework between the feature generator and the classifier.
In particular, the bi-classifier model {MCD}~\cite{saito2018maximum} does the min-max optimization on the discrepancy between classifiers which coincides with the theory of DA~\cite{ben2010theory}.
Following the bi-classifier model, {BCDM}~\cite{li2020bi} uses the {CDD} loss to improve the deterministic classification results.
Note that the success of bi-classifier models depends on accessing the both domain data simultaneously to estimate the domain gap accurately. 
The bi-classifier model is further extended to multiple-classifiers model.
For example, {STAR}~\cite{lu2020stochastic} uses stochastic classifiers sampled from the classifier distribution as a strong domain discriminator.
However, STAR does not provide theoretical analysis on why using multiple classifiers rather than bi-classifier and it still requires to use both domain data during adaptation. 

Source free UDA is more challenging than source available UDA because it can not touch the source domain data during adaptation. 
{SHOT}~\cite{liang2020we} tackles this problem by pre-training its source model with label smoothing, and then exploits the pseudo labels obtained by clustering the target samples for source free UDA.
Several works~\cite{li2020model, yang2021generalized, qiu2021source} following the line of research of pseudo labeling have been proposed recently for source free UDA.
Besides the pseudo label based methods, BAIT~\cite{yang2020casting} extends the bi-classifier model to the source free setting.

Comparing with the above approaches, our work bridges theoretical results on many classifiers model to the source free UDA setting. That is to say, we answer why using many classifiers is better for UDA and the optimal number of classifiers for a specific UDA task.

\section{Domain Adaptation via Many Classifiers}
\label{sec:dmac_theory}
\subsection{Problem setting and motivation}
\label{sec:method}
Given a labeled source domain $D_s=\{X_s,Y_s\}$ and an unlabeled target domain $D_t=\{X_t\}$, we study the problem of source free UDA.
Both the source domain and the target domain share the same label space. Our model is composed of a feature generator $G$ and a set of classifiers $\{C_i\}_{i=1}^k$.

Suppose $k$ is the number of classifiers and $c$ is the number of category.
Figure~\ref{fig:introduction} shows the ratio of disagreement for 3-category problem by using the bi-classifier and the 3-classifier, respectively. 
For the simple 3-category classification task by using the bi-classifier, the ratio of disagreement between the two classifiers in the entire hypothesis space is $P_{c=3}^{k=2}=\frac{3\times 2}{3^2} =\frac{2}{3}$, while the ratio of disagreement for 2-category task is $P_{c=2}^{k=2}=\frac{2 \times 1}{2^2}=\frac{1}{2}$.
On the other hand, if we use 3-classifier model for the 3-category problem shown in Figure~\ref{fig:introduction}, the ratio of disagreement decreases to $P_{c=3}^{k=3}=\frac{3\times 2}{3^3}=\frac{2}{9}$.
It is easy to generalize that the ratio of bi-classifier disagreement for c-category problem is $P_{c}^{k=2}=\frac{c(c-1)}{c^2}=\frac{c-1}{c}$. 
As the number of categories increases, the ratio of disagreement between bi-classifier becomes more and more deterministic since $\lim_{c\to\infty}P_{c}^{k=2}=1$.
In view of domain gap estimation by maximizing the discrepancy between bi-classifier~\cite{ben2010theory, saito2018maximum}, the deterministic disagreement of bi-classifier leads to a loose upper bound of domain gap. This motivates us to explore whether more classifiers help to tighten the upper bound of domain gap estimation. A more interesting observation is that our many-classifier model not only bounds the domain gap tighter but the learned source domain knowledge is maximized in the source free setting. 



\subsection{Theory}
\label{sec:theory}

Due to space limitations, this subsection mainly shows the key lemmas, definitions and theorems. More detailed and complete proofs can be found in the supplementary material.
Based on the theory proposed in ~\cite{ben2010theory} which defines the divergence metric of symmetric difference hypothesis space $\mathcal{H}\Delta\mathcal{H}$ by measuring the disagreement between 2 hypotheses, we generalize the divergence to $\cap \mathcal{H}\Delta\mathcal{H}$ by measuring the disagreement by pairs of hypotheses.

First, we extend the $\mathcal{H}\Delta\mathcal{H}$ space to $\cap \mathcal{H}\Delta\mathcal{H}$ by combining $\mathcal{H}\Delta\mathcal{H}$ in pairs to accommodate our many classifiers model. To facilitate calculation, we introduce the soft version of $\cap \mathcal{H}\Delta\mathcal{H}$ and the induced $d_{\cap\mathcal{H} \Delta\mathcal{H}}$-divergence. 
\begin{definition}[Soft $\cap \mathcal{H}\Delta\mathcal{H}$-space]
\label{def:multiple_hypothesis_space}
For a hypothesis set $H_k\subset \mathcal{H}$, where $H_k$ is a set of $k$ hypotheses, the multiple difference hypothesis space set $\cap \mathcal{H}\Delta\mathcal{H}$ is the intersection of symmetric difference hypotheses space $\mathcal{H}\Delta\mathcal{H}$.
  \begin{equation}
    g\in \cap \mathcal{H}\Delta\mathcal{H}\implies g=\frac{\sum h_i\oplus h_j}{C_k^2},~\forall (h_i,h_j)\in \dbinom{H_k}{2}
  \end{equation}
where $C_k^2$ is the number of pairs of classifiers from $\mathcal{H}_k$. $g$ is defined as the expectation of XOR function for pairs of classifiers.
\end{definition}

\begin{definition}[$d_{\cap\mathcal{H} \Delta\mathcal{H}}$-divergence]\label{def:d_manyclassifiers}
For $\forall g\in \cap \mathcal{H}\Delta\mathcal{H}$ in the Definition~\ref{def:multiple_hypothesis_space} of soft $\cap \mathcal{H}\Delta\mathcal{H}$-space, when there are $k$ hypotheses, the $d_{\cap\mathcal{H} \Delta\mathcal{H}}$-distance is:
  \begin{equation}
    \begin{split}
      d_c^{k}(\mathcal{D}_S,\mathcal{D}_T)=2\sup _{g\in \cap \mathcal{H}\Delta\mathcal{H}}|P_{\mathcal{D}_S}[I(g)]-P_{\mathcal{D}_T}[I(g)]|
    \end{split}
  \end{equation}
\end{definition}
When using many classifiers, the domain gap is the maximization of the expectation that hypotheses are inconsistent across multiple hypotheses. The Definition~\ref{def:d_manyclassifiers} of $d_{\cap\mathcal{H} \Delta\mathcal{H}}$-divergence requires accessing both source and target domain to touch domain boundaries. But for source free UDA, when pre-training in the source domain, the classifiers need to be separated as far as possible. In the ideal case some classifiers need to touch the decision boundary whilst ensuring accurate classification. This placement of classifier guarantees that the gap between the source domain and the unknown target domain can be measured more stable by increasing the number of classifiers.

Next, we will demonstrate the superiority of using more classifiers and the optimal number of classifiers from the disagreement view between multiple hypotheses.
\begin{definition}[Disagreement of hypothesis set $H_k$]
For $\forall h_i,h_j\in H_k$, the disagreement $\epsilon(H_k,H_k)$ is expectation of the mean value of XOR function for
all hypothesis combinations $(h_i,h_j)\in \binom{H_k}{2}$.
  \begin{equation}
      \epsilon(H_k,H_k)=E_{x\sim D} [\frac{\sum h_i\oplus h_j}{C_k^2}] 
  \end{equation}
\end{definition}
Suppose that the number of categories is $c$, when $k\in [2,c]$,
  \begin{equation}
    \dfrac{P_c^{k}}{P_c^{k-1}}=\dfrac{A_c^k/c^k}{A_c^{k-1}/c^{k-1}}=\dfrac{c-k+1}{c}\in [\dfrac{1}{c},\dfrac{c-1}{c}]\le 1
  \end{equation}
where $A^k_c$ denotes the number of $k$-permutations of $c$. Then using more classifiers leads to a smaller ratio of disagreement. Note that if $k > c$, the extra $k-c$ classifiers can not bring more disagreements. For example, if we use 3-classifier model for the 2-category problem, we will have $P_2^3 = P_2^2 = \frac{1}{2}$ because the three classifiers can be grouped into three groups of bi-classifier in which the ratio of disagreement all equals $\frac{1}{2}$. Obviously,  we have $P_c^{c+1} = P_c^c$.

\begin{proposition}
\label{theom:law}
For c-category classification task, if $k<c$, (k+1)-classifier model provides tighter upper bound estimation than k-classifier model. If $k=c$, (k+1)-classifier model provides the same upper bound estimation as k-classifier model.
  \begin{eqnarray}
      {d_c^k}_{\mathcal{H}\Delta\mathcal{H}}(\mathcal{D}_S,\mathcal{D}_T)&=&\dfrac{c-k+1}{c}{d_c^{k-1}}_{\mathcal{H}\Delta\mathcal{H}}(\mathcal{D}_S,\mathcal{D}_T), k\le c\\
      {d_c^k}_{\mathcal{H}\Delta\mathcal{H}}(\mathcal{D}_S,\mathcal{D}_T)&=&{d_c^{k+1}}_{\mathcal{H}\Delta\mathcal{H}}(\mathcal{D}_S,\mathcal{D}_T), k\ge c
  \end{eqnarray}
where \begin{scriptsize}$\dfrac{c-k+1}{c} \in [\dfrac{1}{c},1)$\end{scriptsize}.
\end{proposition}

From the theory of target error bound for the hypothesis space $\mathcal{H}\varDelta \mathcal{H}$~\cite{ben2010theory}, we can derive the target error bound for multiple hypotheses.

\begin{proposition}[Target error bound for $\cap \mathcal{H}\varDelta \mathcal{H}$]
\label{def:tat_error_for_many_cls}
Suppose $\cap \mathcal{H}\varDelta \mathcal{H}$ is the hypothesis space. If $U_S, U_T$ are unlabeled samples from $\mathcal{D}_S$ and $\mathcal{D}_T$, whose size are both $m^\prime$. For $\forall \delta \in (0,1), H_k\subset \mathcal{H}$, which the probability is $1-\delta$ at least:
  \begin{equation}
    \epsilon_t(H_k)\le \epsilon_s(H_k)+\dfrac{1}{2}\hat{d}^{k}_{\cap \mathcal{H}\Delta\mathcal{H}}(\mathcal{U}_S,\mathcal{U}_T)+4\sqrt{\dfrac{2d\log (2m^\prime)+\log (\dfrac{2}{\delta})}{m^\prime}}+\lambda
  \end{equation}
\end{proposition}

From the Proposition~\ref{theom:law} and the Proposition~\ref{def:tat_error_for_many_cls}, we can give the optimal number of classifiers.
\begin{proposition}[Optimal number of classifiers]
\label{theom:the_optimal_number_of_cls}
For the c-category task, the  c-classifier model provides the tightest upper error bound.
\end{proposition}

In practice, the number of categories $c$ can be large. Using the optimal number of classifiers could theoretically bound the upper bound tightest though, the tightest upper bound can not be achieved with limited training data. On the other hand, for limited training data, using too many classifiers would lead to unnecessary computation. This leads us to think about the trade-off of domain gap estimation for the task of many categories. 
For practical source free UDA tasks, if $c$ is large, we propose to use $k$ classifiers which induce the ratio of disagreement  \begin{scriptsize}$P_c^{k}=A_c^k/c^k\approx \epsilon$\end{scriptsize}. Here $\epsilon$ is a predefined threshold of disagreement ratio which controls how tight of the domain gap estimation our model should achieve. We explore the trade-off in Section~\ref{sec:trade-off-exp} on the Office-Home dataset which has 65 categories.

\section{DAMC: a Source Free Framework}
\label{sec:algorithm}


\begin{figure*}[!t]
\begin{center}
\centerline{\includegraphics[width=1\textwidth]{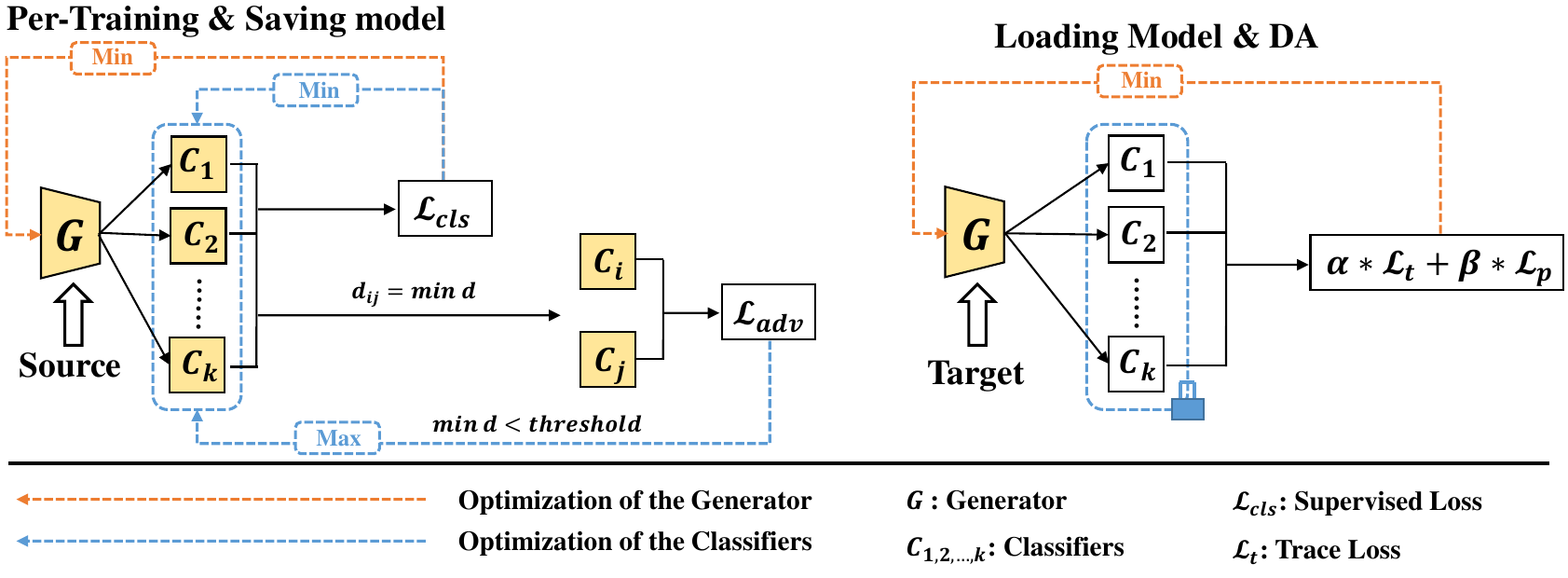}}
\caption{Framework of source free unsupervised domain
adaptation via many classifiers (DAMC). (Left): Source model pre-trains on the source domain data. (Right): Source free target domain adaptation by fixing the pretrained many classifier.}
\label{fig:framework}
\end{center}
\end{figure*}

Based on our theoretical results of the many classifiers model, we propose the framework of source free UDA via many classifiers (DAMC) which consists of two modules shown in Figure~\ref{fig:framework}. The first module is to pre-train the source model with a feature generator $G$ and many classifiers $\{C_i\}_{i=1}^k$ on labeled source domain, which pushes away the decision boundaries of each pair of classifiers whilst ensuring the correctness by supervised learning. The second module is to adapt the source model to the target domain by optimizing the feature generator to maximize the agreements among all classifiers. 

\subsection{Pretraining the source model}

First, the source model is trained by using the labeled source data to learn the mapping $f_S:X_s\rightarrow Y_s$. Note that the source model is composed of a set of classifiers $\{C_i\}_{i=1}^k$ and a feature generator $G$, then following the supervised learning paradigm, the cross entropy loss~(Equation~\eqref{loss:cls}) is used for training $\{C_i\}_{i=1}^k$ and $G$.
  \begin{equation}
    \label{loss:cls}
    \begin{split}
      \min_{G,C}\mathcal{L} _{cls}\left(X_s,Y_s\right) =-\mathbb{E} _{\left(X_s,Y_s\right) }\left[\frac{1}{k}\sum _{i=1}^k y_s\cdot \log \left(C_i\left(G\left(X_s\right) \right) \right) \right]
    \end{split}
  \end{equation}
%
\begin{figure*}[!t]
\vskip 0.2in
\begin{center}
\centerline{\includegraphics[width=0.9\columnwidth]{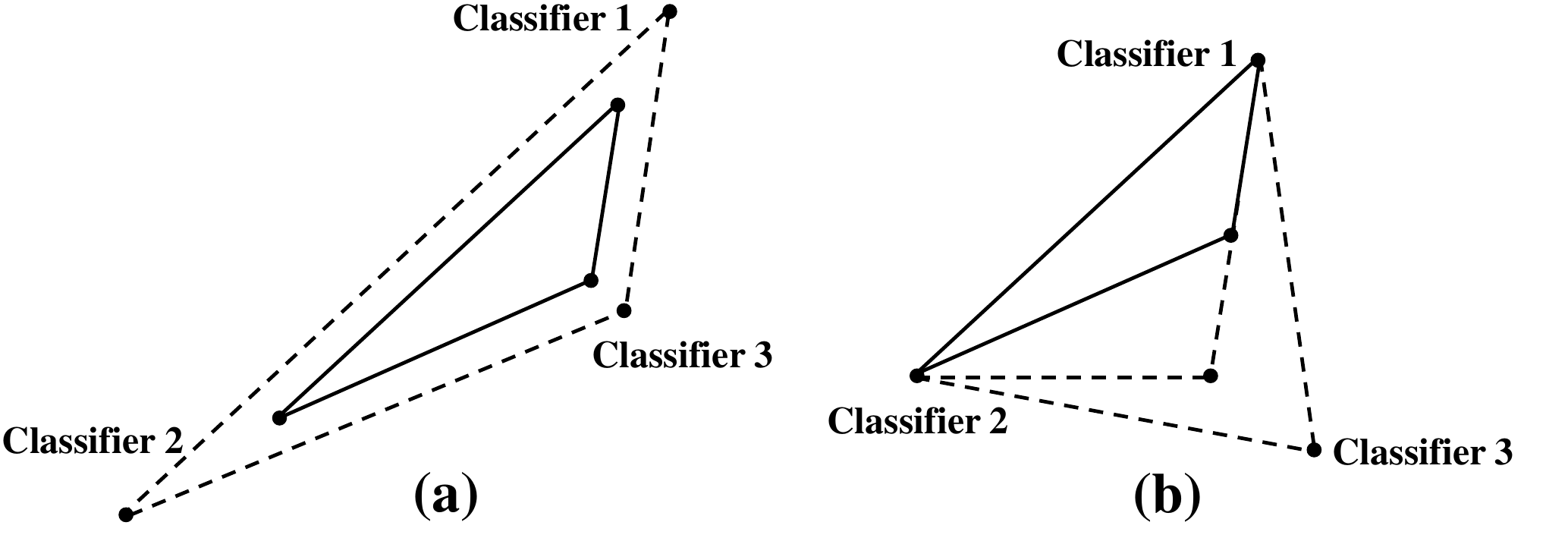}}
\caption{
Comparison between (a) maximizing the accumulated discrepancy among all classifiers and (b) our proposed worst optimization method.}
\label{fig:adv_bost}
\end{center}
\vskip -0.2in
\end{figure*}

Second, based on our theory of maximizing the disagreement among many classifiers, the decision boundaries of classifiers should be placed as far as possible from each other. However, simply maximizing the accumulated discrepancy among all classifiers heavily depends on the initial placements of classifiers. Figure~\ref{fig:adv_bost}
~(a) shows the failure case that classifiers can not be optimized to achieve the maximum disagreements with the same learning rate. 
From geometric view, the best placement of the set of classifiers should be a regular simplex shown in Figure~\ref{fig:adv_bost}~(b), which induces the largest disagreement. We introduce a worst optimization approach to place the classifiers as close as the regular simplex. The loss function is shown in Equation~\eqref{loss:adv}. 
  \begin{equation}
    \label{loss:adv}
    \begin{split}
      \max_C \mathcal{L} _{adv}\left(X_s\right)=\alpha_s *|p_i-p_j|_{\min},~\forall (h_i,h_j)\in \dbinom{H_k}{2}
    \end{split}
  \end{equation}
Here $\alpha_s$ is the coefficient of the adversarial loss for pushing classifiers away.
Figure~\ref{fig:adv_bost}~(b) demonstrates this idea. 
We consider to maximize the closest pair of classifiers until the smallest discrepancy larger than the prescribed threshold $\tau$.  This kind of worst optimization is robust and stable even though some classifiers of our many classifier model are initialized very close. 
Algorithm~\ref{alg:algorithm_pre_train} demonstrates our pre-training strategy.

\begin{algorithm*}[!t]\footnotesize
   \caption{Per-training the Source Model}
   \label{alg:algorithm_pre_train}
\begin{algorithmic}[1]
   \REQUIRE labeled source data $D_s=\{X_s,Y_s\}$, feature generator $G$, a set of classifiers $C_1,C_2,...,C_k$, prescribed threshold $\tau$.
   \STATE Random initialization the parameters of $G$ and $C_1,C_2,...,C_k$.
   \STATE $p=0,t=0$
   \REPEAT
   \STATE Sample $\{X_{s_i},Y_{s_i}\}$ from labeled source data.
   \STATE Update $C_1,C_2...,C_k$ and $G$ by minimizing the cross entropy loss on $D_s=\{X_s,Y_s\}$ according to Equation~\eqref{loss:cls}.
   \STATE Extract the source feature: $feat_s = G(X_s)$
   \REPEAT 
   \STATE Softmax output for each classifier: \\$p_i = Softmax(C_i(feat_s))$
   \STATE Fix $G$ and update the closest pair of $C_i^*$ and $C_j^*$ by maximizing Equation~\eqref{loss:adv}.
   \UNTIL{$\min{\{ |p_i - p_j| \} }\geq \tau$ }
   \STATE $p=p+1$
   \UNTIL{$p>=max\_src\_epoch$}
   \ENSURE $C_1,C_2,...,C_k$ and $G$.
\end{algorithmic}
\end{algorithm*}

%
\subsection{Source free target domain adaptation}

\begin{algorithm}[tb]\footnotesize
   \caption{Domain Adaptation for the Target Domain}
   \label{alg:algorithm_domain_adaptation}
\begin{algorithmic}[1]
   \REQUIRE unlabeled target data $D_t=\{X_t\}$, pre-trained feature generator $G$, pre-trained classifiers $C_1,C_2...,C_k$
   \STATE Load the parameters of the pre-trained $G$ and the pre-trained $C_1,C_2...,C_k$.
   \STATE $p=0$
   \REPEAT
   \STATE Get the pseudo-label of the target domain $\hat{y_t}$.
   \STATE Sample $\{X_{t_i}\}$ from unlabeled target data.
   \STATE  Fix $C_1,C_2,...,C_k$, and update $G$ by minimizing Equation~\eqref{loss:DA}.
   \STATE $p=p+1$
   \UNTIL{$p>=max\_tgt\_epoch$}
   \ENSURE the posterior distribution $p(y|x_i)$ of the target
\end{algorithmic}
\end{algorithm}
Once the source model is pretrained, the set of classifiers $\{C_i\}_{i=1}^k$ are placed as far as possible whilst ensuring the correctness in the source domain. According to the theory discussed in Section~\ref{sec:theory}, $\{C_i\}_{i=1}^k$ provides a much tighter upper bound of domain gap in the source free setting. Since we can not touch source data in adaption, the set of classifiers $\{C_i\}_{i=1}^k$ should be frozen to act as a metric of estimating the disagreement among classifiers. Then for target adaptation, we will fine-tune the pretrained $G$ to maximize the agreements among all classifiers.
Ideally, Equation~\eqref{loss:tr} measures this disagreement exactly where $p_i$ is the softmax output of each classifier $C_i$. 
  \begin{equation}
    \label{loss:tr}
    \begin{split}
      \mathcal{L}_{tr}(X_t)=1-\sum_{i=1}^c \prod_{j=1}^k p_i^j
    \end{split}
  \end{equation}
Note that $\sum_{i=1}^c \prod_{j=1}^k p_i^j$ defines the agreement among classifiers which corresponding to the trace of a k-dimensional tensor composed of $\{p_1, p_2, \ldots, p_k\}$

However, the trace loss~\eqref{loss:tr} is highly non-convex and is difficult to be optimized. Thus, we propose to use a soft version to surrogate the trace loss~\eqref{loss:tr}. We consider the trace loss of each pair of classifiers with all possible combination. Specifically, for the set of k classifiers $H_k$, we will sum up {\tiny $\dbinom{H_k}{2}$}
pairs of loss, referring to Equation~\eqref{loss:pair_tr}.
Note that the trace loss of each pair of classifiers coincides with the CDD loss used in BCDM~\cite{li2020bi}. However we propose this soft version of metric in terms of disagreement, whereas BCDM aims to improve the classification certainty.
  \begin{equation}
    \label{loss:pair_tr}
    \begin{split}
      \mathcal{L}_{pair\_tr}(X_t)=\sum(1-\sum_{i=1}^c p_i q_i),~\forall (p,q)\in \dbinom{H_k}{2}
    \end{split}
  \end{equation}

Furthermore, followed by SHOT~\cite{liang2020we}, we also use conditional entropy {$\mathcal{L}_{ent}^c$}, marginal entropy { $\mathcal{L}_{ent}^m$}, and pseudo-label loss { $\mathcal{L}_{p}$} as regularization terms during target adaptation. Finally, our optimization goal is:
  \begin{equation}
  \label{loss:DA}
    \begin{split}
      \min \alpha_t*\mathcal{L}_{pair\_tr}(X_t)+ \gamma_1*\mathcal{L}_{ent}^c - \gamma_2*\mathcal{L}_{ent}^m + \beta*\mathcal{L}_{p}
    \end{split}
  \end{equation}
In this work the weights of both entropy losses $\gamma_1$ and $\gamma_2$ are always set 0.1.
$\alpha_t$ is the coefficient of the trace loss and $\beta$ is the coefficient of pseudo-label loss.
Algorithm~\ref{alg:algorithm_domain_adaptation} describes the target domain adaptation of our DAMC. The target adaptation loss~\eqref{loss:DA} consists pseudo-label loss though, we emphasize that it does help to correct our model from global view in our DAMC framework, the overall performance is dominated by our { $\mathcal{L}_{pair\_tr}(X_t)$ }loss. We investigate the two loss terms in ablation study referring to Section~\ref{subsubsec:pseudo-trace}.

\subsection{Model selection}
\label{sec:model_sel}
Since the target domain data cannot be accessed during pre-training, our many classifiers model requires to place the boundaries of classifiers as far as possible whilst ensuring accurate classification to improve the possibility of disagreement among classifiers in the target domain. Note that all classifiers reaching 100\% training accuracy does not provide the maximum disagreement in the source domain, not to mention the target domain. On the other hand, if all classifiers can not reach 100\% training accuracy simultaneously, it means that all classifiers lie on the decision boundary which can also be degenerated to the case that all classifiers only disagree on one category.

To tackle this, we use a heuristic model selection strategy as follows. If a classifier can not reach 100\% training accuracy on a category, it means that the classifier is placed on the decision boundary of this category. To assure the correctness, we also require more classifiers to reach 100\% training accuracy on this category. That is to say, when training our source model, we will record the training accuracy of all classifiers for each category. The ideal source model is that for each category only 1-2 classifiers touch the decision boundary and all other classifiers predict 100\% accuracy. This model selection strategy is particular useful for the dataset with enough training data such as VisDA.

\section{Experiment}

\label{sec:experiments}
\subsection{Datasets and implementation details}
\label{sub:datasets}
\textbf{Datasets.}
To verify the performance of our introduced new method, we select commonly used public domain adaptation datasets for experiments.
We performed ablation study on \textbf{VisDA}~\cite{peng2017visda} to verify the effects of many of our proposed classifier models from various angles.
\textbf{VisDA}\label{dataset:visda} represents the largest cross-domain object classification, with over 280K images in 12 categories combining training, validation, and test domains.
Furthermore, we also test the performance of our algorithm on the simple dataset \textbf{Digits} and the medium-scale dataset \textbf{Office-Home}~\cite{venkateswara2017deep}.
\textbf{Digits} are used to evaluate three types of adaptation scenarios, 10 categories ranging from 0 to 9, including three datasets: SVHN(S), MNIST(M), and USPS(U).
SVHN is the Street View Door Numbers dataset, which consists of cropped color digital images from real Street View.
MNIST and USPS are datasets of handwritten digits.
Office-Home\label{dataset:officehome} is a commonly used benchmark dataset in the Domain Adaptation field, with 65 categories and four data domains: Art(A), Clip(C), Product(P) and RealWorld (R).

\textbf{Network Architecture.}
In this work, for digits tasks, we adopt the CNN architecture used in \cite{ganin2015unsupervised} and \cite{bousmalis2017unsupervised} for feature generation, and use fully connected layers as classifiers.
For medium-scale and large-scale benchmarks, we adopt ResNet~\cite{he2016deep} as pre-trained base models for VisDA (ResNet-101) and Office-Home tasks (ResNet-50).
Except the last fully connected layer, ResNet is used as a feature generator. Bottleneck layer is used to squeeze the feature dimension to 256. For classifiers, we use three fully connected layers for VisDA and two fully connected layers for Office-Home.
Batch normalization~\cite{pmlr-v37-ioffe15} layers are inserted into the classifiers for each task. Batch size is set to 32. Our models are optimized by SGD with momentum\cite{sutskever2013importance} and the learning rate is scheduled by the same scheduler as~\cite{ganin2015unsupervised}.
\begin{equation}
    \eta = \eta_0 \cdot \left(1 + \frac{10\cdot iter}{maxIter}\right)^{-0.75}
\end{equation}


Detailed network architecture and hyper-parameters including learning rate settings and network structures for Digits, VisDA and Office-Home can be found in the Appendix.

\begin{table}\footnotesize
\caption{Adaptation accuracy(\%) on the UDA task {Digits}.}
\vskip -0.1in
  \label{Experiment:digital}
    \begin{center}
    \scalebox{0.7}{
    \begin{tabular}{c|c|c c c|c}
      \hline
      Method(Src-Tag) & S.F. & S$\rightarrow$M & M$\rightarrow$U & U$\rightarrow$M & Avg.\\
      \hline
      Source Only & - & 67.1 & 76.7 & 63.4 & 69.1\\
      ADDA\cite{tzeng2017adversarial} & - & 76.0 & 89.4 & 90.1 & 85.2\\
      ADR\cite{saito2018adversarial} & - & 94.1 & 91.3 & 91.5 & 92.3\\
      MCD\cite{saito2018maximum} & - & 96.2 & 94.2 & 94.1 & 94.8\\
      3CATN\cite{li2019cycle} & - & 92.5 & 96.1 & 98.3 & 95.6\\
      SWD\cite{lee2019sliced} & - & 98.9 & 98.1 & 97.1 & 98.0\\
      DM-ADA\cite{xu2019adversarial} & - & 95.5 & 96.7 & 94.2 & 95.5\\
      CycleGAN\cite{ye2020light} & - & 97.5 & 97.1 & 98.3 & 97.6\\
      STAR\cite{lu2020stochastic} & - & 98.8 & 97.8 & 97.7 & 98.1\\
      \hline
      SHOT\cite{liang2020we} & \checkmark & 98.9 & 98.4 & 98.0 & 98.4\\
      Ours(k=10) & \checkmark & \textbf{99.2} & \textbf{98.4} & \textbf{98.6} & \textbf{98.7}\\
      \hline
    \end{tabular}
    }
    \end{center}
\end{table} 

\subsection{Results of public data sets}
\label{subsec:Results of all}

%
%
%

\subsubsection{Digits}

For the Digits dataset, since there are 10 categories, our model consist of 10 classifiers $\{C_i\}_i^{k=10}$ and a feature generator $G$. We evaluated the classification accuracy of DAMC on three cross-domain tasks of digits for source free UDA.
We did 20 epochs source model pre-training and selected the optimal model according to Section~\ref{sec:model_sel}. For target adaptation, ``SVHN(S) to MNIST(M)" required 100 epochs. Both ``MNIST(M) to USPS(U)" and ``USPS(U) to MNIST(M)" required 500 epochs.
The experimental results are shown in Table~\ref{Experiment:digital}.
DAMC achieves the best classification accuracy on all three tasks including source available UDA approaches. 

\subsubsection{VisDA}

\begin{table*}[t] 
\caption{Adaptation accuracy(\%) on the UDA task \textbf{VisDA} (backbone ResNet-101). }
\vskip -0.3in
  \label{Experiment:visDA}
  \begin{center}
  \scalebox{0.6}
  {
    \begin{tabular}{c|c|c c c c c c c c c c c c|c}
      \hline
         Method & S.F. & plane & bcycl & bus & car & horse & knife & mcycl & person & plant & sktbrd & train & truck & Avg.\\
      \hline
      Source Only & - & 60.9 & 21.6 & 50.9  & 67.7 & 65.8 & 6.3 & 82.2 & 23.2 & 57.3 & 30.6 & 84.6 & 8.0 & 46.6\\
      ResNet-101\cite{he2016deep} & - & 55.1 & 53.3 & 61.9 &59.1 & 80.6 & 17.9 & 79.7 & 31.2 & 81.0 & 26.5 & 73.5 & 8.5 & 52.4\\
      MCD\cite{saito2018maximum} & - & 87.0 & 60.9 & 83.7 & 64.0 & 88.9 & 79.6 & 84.7 & 76.9 & 88.6 & 40.3 & 83.0 & 25.8 & 71.9\\
      SAFN\cite{xu2019larger} & - & 93.6 & 61.3 & 84.1 & 70.6 & 94.1 & 79.0 & 91.8 & 79.6 & 89.9 & 55.6 & 89.0 & 24.4 & 76.1 \\
      JADA\cite{li2019joint} & - & 91.9 & 78.0 & 81.5 & 68.7 & 90.2 & 84.1 & 84.0 & 73.6 & 88.2 & 67.2 & 79.0 & 38.0 & 77.0\\
      SWD\cite{lee2019sliced} & - & 90.8 & 82.5 & 81.7 & 70.5 & 91.7 & 69.5 & 86.3 & 77.5 & 87.4 & 63.6 & 85.6 & 26.2 & 76.4\\
      BCDM\cite{li2020bi} & - & 95.1 & 87.6 & 81.2 & 73.2 & 92.7 & 95.4 & 86.9 & 82.5 & 95.1 & 84.8 & 88.1 & 39.5 & 83.4\\
      STAR\cite{lu2020stochastic} & - & 95.0 & 84.0 & 84.6 & 73.0 & 91.6 & 91.8 & 85.9 & 78.4 & 94.4 & 84.7 & 87.0 & 42.2 & 82.7\\
      \hline
      SHOT\cite{liang2020we} & \checkmark & 94.3 & 88.5 & 80.1 & 57.3 & 93.1 & 94.9 & 80.7 & 80.3 & 91.5 & 89.1 & 86.3 & 58.2 & 82.9\\
      G-SFDA\cite{yang2021generalized} &\checkmark &\textbf{96.1} &88.3 &85.5 &74.1 &\textbf{97.1} &95.4 &89.5 &79.4 &95.4 & 92.9 &\textbf{89.1} &42.6 &85.4\\
      CPGA(40ep)\cite{qiu2021source}& \checkmark &94.8 &83.6 &79.7 &65.1 &92.5 &94.7 &90.1 &82.4 &88.8 &88.0 &88.9 &60.1 &84.1\\
      CPGA(400ep)\cite{qiu2021source}& \checkmark &95.6 &\textbf{89.0} &75.4 &64.9 &91.7 &\textbf{97.5} &89.7 &83.8 &93.9 &\textbf{93.4} &87.7 &\textbf{69.0} &\textbf{86.0}\\
      \hline
      Ours(k=12) & \checkmark & 95.4 & 86.9 & \textbf{87.2} & \textbf{79.7} & 94.6 & 95.7 & \textbf{91.2} & \textbf{84.9} & \textbf{95.7} & 91.5 & 81.9 & 41.7 & \textbf{86.0}\\
      \hline
    \end{tabular}
    }
  \end{center}
\end{table*} 

For VisDA, we used a 12-classifier model since there are 12 categories. We pretrained the source model for 20 epochs and selected the optimal model according to Section~\ref{sec:model_sel}. Then performed 30 epochs for target adaptation.
Note that, the pseudo label loss was applied starting from the second epoch and pseudo labels were re-calculated in every two epochs. Table~\ref{Experiment:visDA} demonstrates the comparison between our DAMC and other state of the art approaches. In particular, compared with other source free methods, DAMC achieves the best overall performance. Note that CPGA~\cite{qiu2021source} requires 400 epochs to achieve this transferring accuracy but DAMC only requires less than 30 epochs. 

Compared with the other methods for source-free, it can be observed that the improvement of DAMC is robust and balanced on almost every category. The reason is twofold. First, the best 12-classifier model is selected based on the model selection strategy discussed in Section~\ref{sec:model_sel}. The 12 classifiers not only guarantee the correctness for each category but also touch the decision boundaries to achieve the largest disagreement. Second, we leverage the pseudo label loss as regularization to incorporate global distribution information of the target domain. 
More study on VisDA can be found in our ablation study~\ref{subsubsec:pseudo-trace}. 
\subsubsection{Office-Home}

\begin{table*}[t] 
\caption{Adaptation accuracy(\%) on the UDA task \textbf{Office-Home} (backbone ResNet-50).}
\vskip -0.1in
\renewcommand\tabcolsep{3.5pt}
  \label{Experiment:Office-Home}
  \begin{center}
  \scalebox{0.65}
  {
    \begin{tabular}{c|c|c c c c c c c c c c c c|c}
      \hline
      Method & S.F. & A$\rightarrow$C & A$\rightarrow$P & A$\rightarrow$R & C$\rightarrow$A & C$\rightarrow$P & C$\rightarrow$R & P$\rightarrow$A & P$\rightarrow$C & P$\rightarrow$R & R$\rightarrow$A & R$\rightarrow$C & R$\rightarrow$P & Avg.\\
      \hline
      ResNet-50\cite{he2016deep} & - & 34.9 & 50.0 & 58.0 & 37.4 & 41.9 & 46.2 & 38.5 & 31.2 & 60.4 & 53.9 & 41.2 & 59.9 & 46.1\\
      DAN\cite{long2015learning} & - & 43.6 & 57.0 & 67.9 & 45.8 & 56.5 & 60.4 & 44.0 & 43.6 & 67.7 & 63.1 & 51.5 & 74.3 & 56.3\\
      SAFN\cite{xu2019larger} & - & 52.0 & 71.7 & 76.3 & 64.2 & 69.9 & 71.9 & 63.7 & 51.4 & 77.1 & 70.9 & 57.1 & 81.5 & 67.3\\
      BDG\cite{yang2020bidirectional} & - &51.5 &73.4 &78.7 &65.3 &71.5 &73.7 &65.1 &49.7 &81.1 &74.6 &55.1 &84.8 &68.7\\
      SRDC\cite{tang2020unsupervised} & - &52.3 &76.3 &81.0 &\textbf{69.5} &76.2 &78.0 &\textbf{68.7} &53.8 &81.7 &\textbf{76.3} &57.1 &\textbf{85.0} &71.3\\
      DFA\cite{wang2021discriminative} & - & 50.6 & 74.8 & 79.3 & 65.2 & 73.8 & 74.5 & 63.5 & 51.4 & 81.4 & 73.9 & 58.2 & 83.3 & 69.2\\
      \hline
      SHOT\cite{liang2020we} & \checkmark & 57.1 & \textbf{78.7} & 81.5 & 68.0 & \textbf{78.2} & \textbf{78.1} & \underline{67.4} & 54.9 & \textbf{82.2} & 73.3 & 58.8 & 84.3 & \textbf{71.8}\\
      BAIT\cite{yang2020casting}& \checkmark &57.4 &77.5 &\textbf{82.4} & 68.0 &77.2 &75.1 &67.1 &55.5 &81.9 &\underline{73.9} &\textbf{59.5} &84.2 &71.6\\
      G-SFDA\cite{yang2021generalized} &\checkmark & 57.9 &78.6 &81.0 &66.7 &77.2 &77.2 &65.6 &\textbf{56.0} &\textbf{82.2} &72.0 &57.8 &83.4 &71.3\\
      \hline
      Ours(k=65) & \checkmark & \textbf{57.9} & 78.5 & 81.1 & 66.7 & 77.7 & 77.9 & 66.5 & 54.3 & 81.5 & 73.2 & 58.9& \textbf{\underline{84.7}} &71.6\\
      \hline
    \end{tabular}
    }
  \end{center}
\end{table*} 

We use a 65-classifier model for Office-Home.
Unlike VisDA, Office-Home does not have enough training samples on both domain.
For example, the Art domain only has 2427 images in total but has 65 categories.
The least category only has 15 images. In this setting, training many classifiers by placing the decision boundaries as far as possible whilst ensuring the correctness becomes more difficult.
However, even this, Table~\ref{Experiment:Office-Home} shows that our DAMC still can compete other UDA approaches by the average transferring accuracy of 71.6.
Note that we did 200 epochs for pre-training the source domain, selected the optimal model according to Section~\ref{sec:model_sel}, and performed 30 epochs transferring for the target domain.
In addition, the pseudo label was used starting from the first epoch and was calculated in every two epochs.

\subsubsection{Online Source Free Domain Adaptation}

\begin{table*}[!t] 
\caption{Adaptation accuracy(\%) of VisDA and Office-Home for Online UDA}
\vskip -0.3in
  \label{online}
  \begin{center}
  \scalebox{0.63}
  {
    \begin{tabular}{c|c c c c c c c c c c c c|c}
      \hline
      VisDA & plane & bcycl & bus & car & horse & knife & mcycl & person & plant & sktbrd & train & truck & Avg.\\
      \hline
      BAIT\cite{yang2020casting}  &93.2& 66.2& \textbf{87.1} &59.1 &90.2& 76.9& 92.1& 83.4& 80.6 &49.5& \textbf{87.7}&\textbf{45.7}  &76.0\\
      \hline
      Ours(1st-ep) &\textbf{94.8}&\textbf{70.0}&86.2&\textbf{73.2}&\textbf{92.2}&\textbf{79.8}&\textbf{93.2}&\textbf{84.9}&\textbf{91.3}&\textbf{71.5}&82.1&36.3&\textbf{79.6}\\
      \hline
      \hline
      Office-Home & A$\rightarrow$C & A$\rightarrow$P & A$\rightarrow$R & C$\rightarrow$A & C$\rightarrow$P & C$\rightarrow$R & P$\rightarrow$A & P$\rightarrow$C & P$\rightarrow$R & R$\rightarrow$A & R$\rightarrow$C & R$\rightarrow$P & Avg.\\
      \hline
      BAIT\cite{yang2020casting}  &51.0& 72.9& 77.4& 60.9 &71.0& 68.7 &\textbf{60.7} &49.3 &\textbf{78.2}& 70.2& 54.5& 80.4 &66.3\\
      \hline
      Ours(1st-ep) & \textbf{53.4}&	\textbf{75.6}&	\textbf{78.6}&	\textbf{63.5}&	\textbf{72.6}&\textbf{73.1}&	59.4&	\textbf{49.6}&76.8&\textbf{70.5}&	\textbf{55.7}&	\textbf{81.1}&	\textbf{67.5}\\
      \hline
    \end{tabular}
    }
  \end{center}
\end{table*} 

Since our model can provide tighter domain gap estimation, we expect it can converge very quickly in adaptation phase. In particular, we examine the performance of adapting the target data only in one epoch, which has been investigated as online source free DA by BAIT~\cite{yang2020casting}. Table~\ref{online} demonstrates the comparison between our model and BAIT on VisDA and Office-Home benchmark tasks. It is a very promising result that our model can achieve average adaptation accuracy of 79.6 on VisDA only in one epoch. Our mode improves $3.6\%$ on VisDA and $1.2\%$ on Office-Home over BAIT. This result confirms that many-classifier model does provide much tighter upper bound of domain gap in source free setting than bi-classifier model which BAIT is. The tighter the domain gap is estimated the faster the target data can be adapted even though only in one epoch. Note that all the hyper-parameters of our model are the same as our prior benchmark experiments on the two datasets, except that the target adaptation epoch is set to 1.

\subsection{Ablation study}
The following ablation studies are mainly conducted on VisDA except the trade-off study of domain gap estimation for the task of many categories in which we use Office-Home.
\subsubsection{Bi-classifier v.s. Many-classifier}
\begin{figure}[!t]
\begin{center}
\centerline{
\renewcommand\tabcolsep{0.1pt}
\begin{tabular}{ccc}
\includegraphics[width=0.5\textwidth]{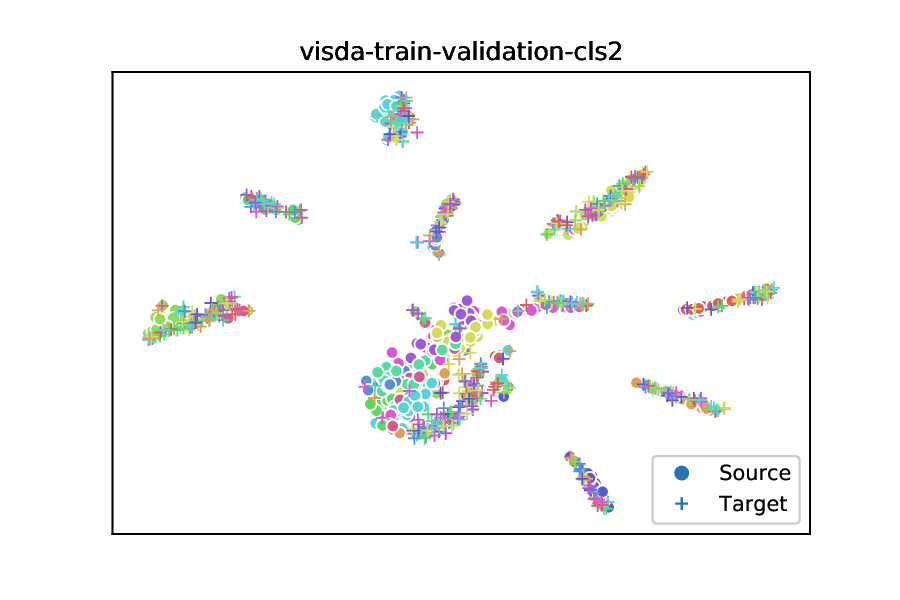}
&
&
\includegraphics[width=0.5\textwidth]{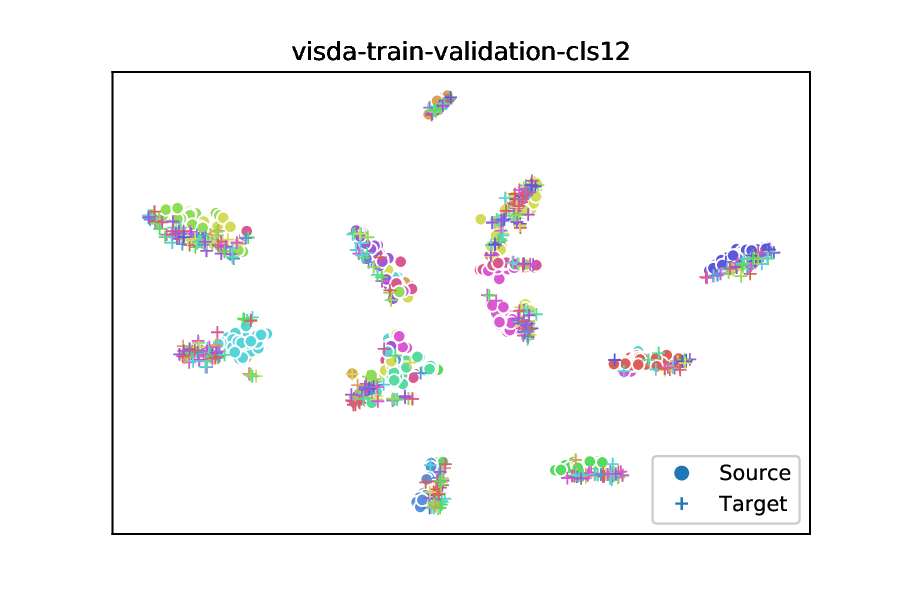}
\\
\end{tabular}}
\vskip -0.2in
\caption{T-SNE visualizations of bi-classifier and 12-classifier. $o$ represents the source domain, $+$ represents the target domain, and different colors represent different categories.}
\label{fig:study1}
\end{center}
\vskip -0.2in
\end{figure}

In this study, we compare our many-classifier model with bi-classifier model to validate that many classifiers do provide tighter domain gap estimation for source free UDA than two classifiers.
Because VisDA has 12 categories, then the best number of classifiers is $k=12$ according to our theory.
We also investigate a 6-classifier model which induces $P^{k=6}_{c=12}=0.223$ ratio of disagreement.
Except excluding the pseudo label loss in Equation~\eqref{loss:DA}, this study follows the same hyper-parameter setting as the previous benchmark experiment on VisDA.
Table~\ref{study-1:visDA} demonstrates the accuracy of each category among these three models at epoch 30.
It is clear that both 6-classifier and 12-classifier outperform almost every category comparing with bi-classifier in VisDA.
To visualize the difference between many-classifier model and bi-classifier model, Figure~\ref{fig:study1} plots the t-SNE embedding~\cite{van2008visualizing} of the features extracted by each $G$.
One is trained along with 12-classifier model, and the other is trained along with bi-classifier model. The 12-classifier model clearly demonstrates better clustering results for the 12 categories than the bi-classifier model.


\begin{table*}[t] 
\caption{Accuracy(\%) of different numbers of classifiers on VisDA.}
\vskip -0.1in
  \label{study-1:visDA}
  \begin{center}
  \scalebox{0.65}{
    \begin{tabular}{c|c c c c c c c c c c c c|c}
      \hline
      Model & plane & bcycl & bus & car & horse & knife & mcycl & person & plant & sktbrd & train & truck & Avg.\\
      \hline
      2-classifier & 94.0 & 85.2 & 85.5 & 73.2 & 92.0 & 98.0 & 84.5 & 77.6 & 89.9 & 77.5 & 84.9 & 28.3 &80.9\\
      6-classifier & 95.3 & 86.5 & 89.1 & 75.0 & 94.5 & 98.2 & 87.3 & 81.2 & 90.9 & 86.8 & 81.0 & 30.4 & 83.0\\
      12-classifier & 94.3 & 86.4 & 87.4 & 77.8 & 93.9 & 98.4 & 89.4 & 80.5 & 93.3 & 86.1 & 83.4 & 49.4 &85.0\\
      \hline
    \end{tabular}
    }
  \end{center}
\end{table*} 

\subsubsection{Trade-off of domain gap estimation for the task of many categories}
\label{sec:trade-off-exp}
Our theory provides the optimal number of classifiers which leads to the tightest upper bound of domain gap estimation though, in practical applications we do encounter the tasks that have many categories, for example the 65-category Office-Home dataset. Ideally, we assume that the source data are sufficient for training. Tighter domain gap estimation requires more classifiers in the source free setting and in the sequel more source data are required to make these classifiers to be trained ``good and far apart". However, the theoretical ratio of disagreement of many classifiers can not be achieved by limited source data. 

\begin{figure*}[!t]
\begin{center}
\centerline{\includegraphics[width=0.8\columnwidth]{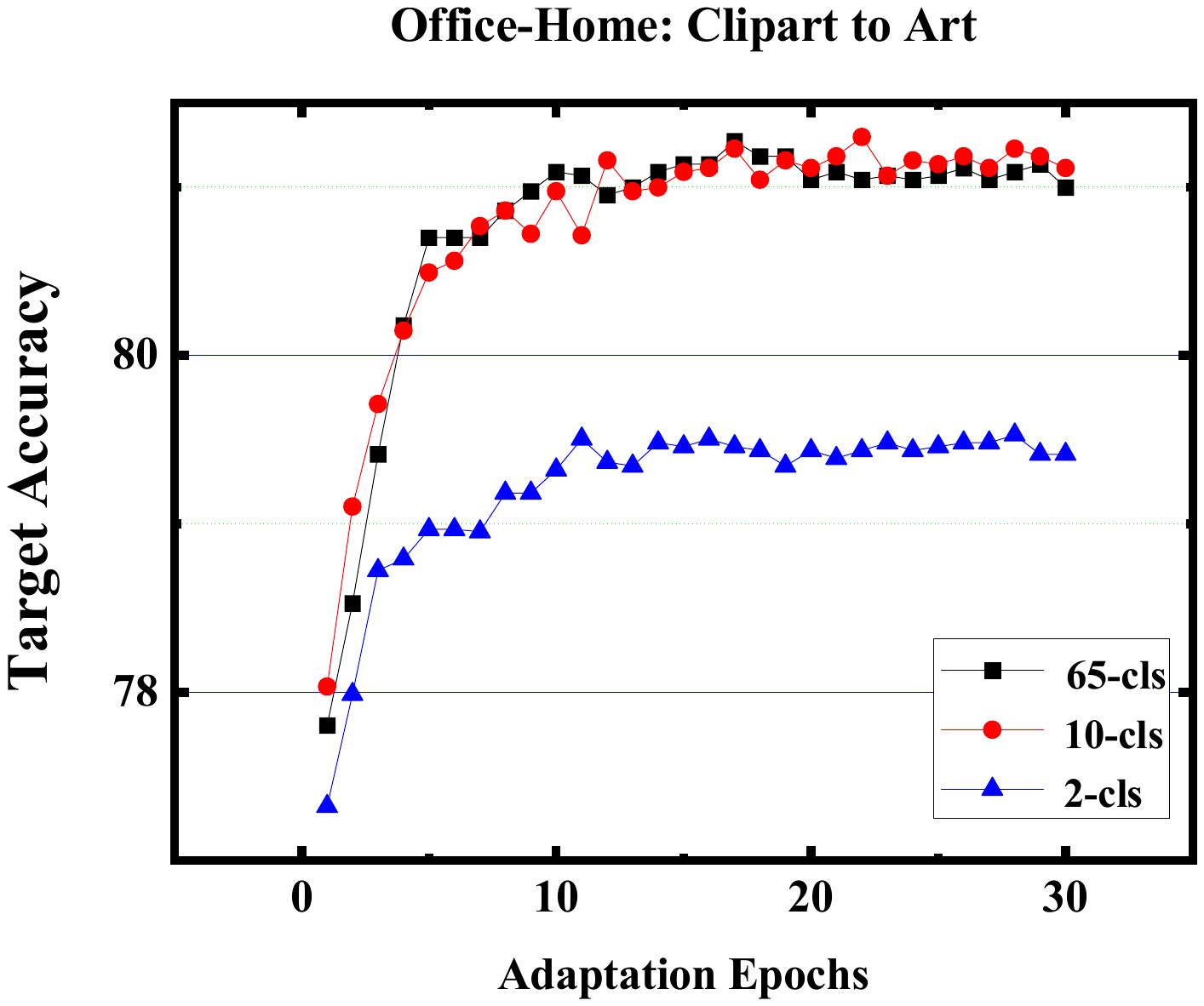}}
\caption{Comparison among the three models of different classifiers for the 65-category Office-Home dataset. For this many categories but limited training data source free UDA task Clipart-Art, 10-classifier model demonstrates the competitive performance of the optimal 65-classifier model.
}
\label{fig:study_trade_off}
\end{center}
\vskip -0.2in
\end{figure*}

In this ablation study, we empirically investigate this trade-off by examining how tight we can achieve for limited source training data. We use Office-Home dataset and take the source free UDA task of Clipart-Art for demonstration. The two domains, Clipart and Art, have 4365 images and 2427 images in total, and have 65 images and 37 images in average,  respectively. For the bi-classifier baseline model,  it can induce $64/65=0.985$ ratio of disagreement. For the optimal 65-classifier model, 
the ideal ratio of disagreement is 1.2E-27 which is impossible to achieve 
from the limited training data. We also choose to use a 10-classifier model which has 0.483 ratio of disagreement. Figure~\ref{fig:study_trade_off} demonstrates the adaptation accuracy of Clipart-Art among these three models at epoch 30. We observe that 10-classifier model is very close to 65-classifier model. 10-classifier model and 65-classifier model clearly outperform the baseline bi-classifier model by a large margin, which validates our theory that tighter domain gap estimation does improve the adaptation accuracy. On the other hand, the close performance between 10-classifier model and the optimal 65-classifier model demonstrates that we won't pursue the tightest domain gap estimation which is impossible for limited training data.

Note that for synthetic images source domain, such as VisDA or autonomous driving simulation scenarios, we do have sufficient synthetic source training data, and thus approaching the tightest domain gap could make the downstream target adaptation have better performance.

%
\subsubsection{Pair of Trace Loss v.s. Pseudo Label Loss}
\label{subsubsec:pseudo-trace}
\begin{figure}[!h]
\begin{center}
\centerline{\includegraphics[width=0.8\columnwidth]{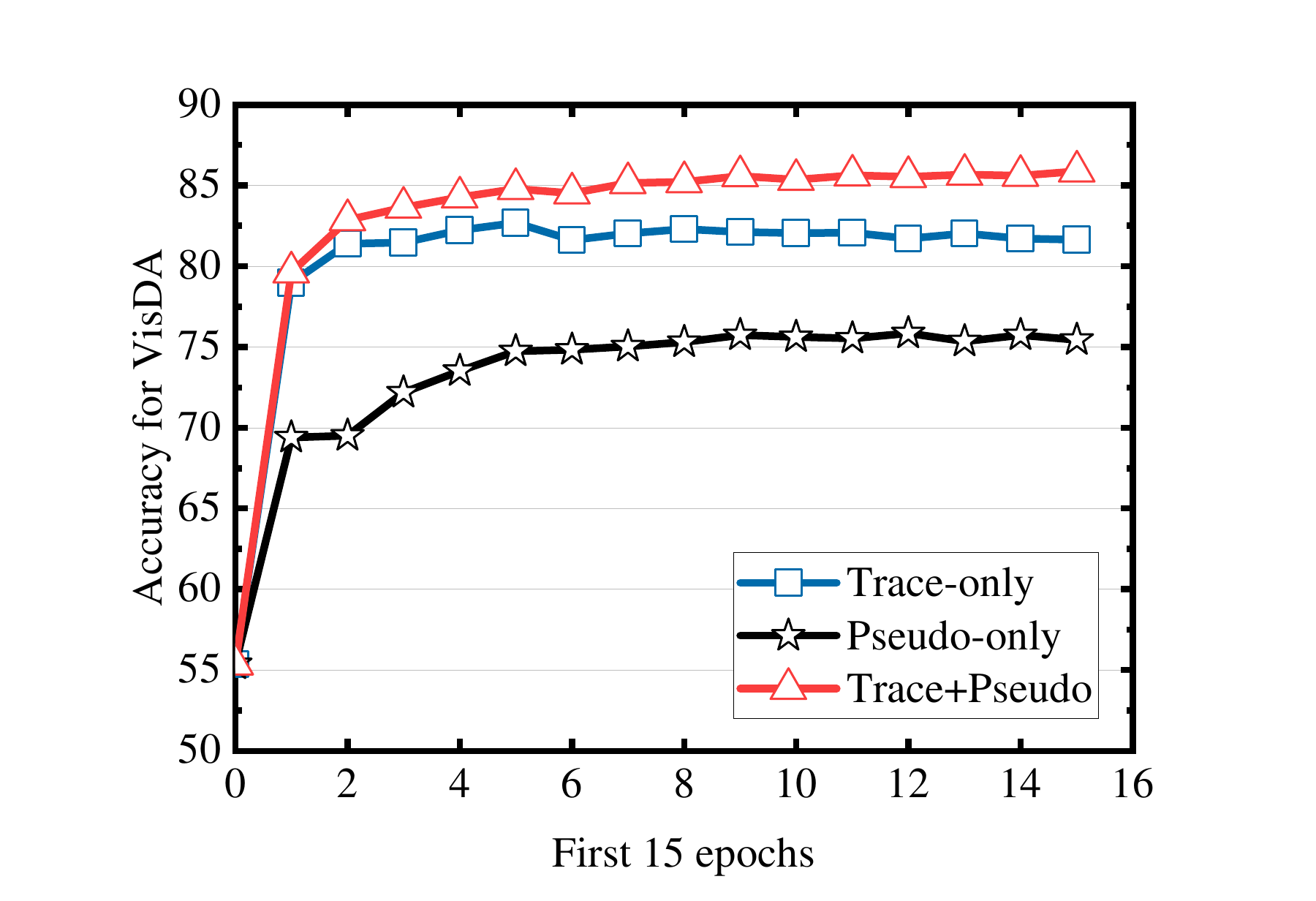}}
\caption{Accuracy plot of ``trace-only", ``pseudo-only" and ``trace \& pseudo" in the first 15 epochs of the VisDA task.}
\label{fig:study_4}
\end{center}
\vskip -0.2in
\end{figure}

\begin{table*}[t] 
\caption{Accuracy(\%) of ``trace-only", ``pseudo-only", and ``trace \& pseudo" on {VisDA}.} 
\vskip -0.3in
  \label{Experiment:study_4}
  \begin{center}
    \scalebox{0.65}{
    \begin{tabular}{c|c c c c c c c c c c c c|c}
      \hline
      Acc(15th-ep) & plane & bcycl & bus & car & horse & knife & mcycl & person & plant & sktbrd & train & truck & Avg.\\
      \hline
      pseudo-only & 94.3 & 81.8 & 81.1 & 64.7 & 90.1 & 27.6 & 83.3 & 76.4 & 86.2 & 88.7 & 85.5 & 45.6 & 75.4\\
      trace-only & 93.6 & 65.8& 85.4 & 76.7 & 90.1 & 95.8 & 95.3 & 83.3 & 92.5 & 85.6 & 84.5 &31.4 & 81.7\\
      trace \& pseudo & 95.3 & 85.8 & 86.7 & 79.4 & 93.7 & 95.8 & 91.4 & 89.9 & 94.9 & 91.8 & 88.3 & 43.5 &85.9\\
      \hline
    \end{tabular}
    }
  \end{center}
\end{table*} 

The target adaptation loss, referring to Equation~\eqref{loss:DA}, uses pseudo-label loss as regularization though, we emphasize that the overall performance is dominated by our pair of trace loss { $\mathcal{L}_{pair\_tr}(X_t)$ }. 
We investigate the two losses in this ablation study.
We compared the results of the first 15 epochs of VisDA by using ``trace-only", ``pseudo-only" and ``trace \& pseudo". ``Trace-only" means that the pseudo label loss is excluded from Equation~\eqref{loss:DA}. ``Pseudo-only" means that the pair of trace loss is excluded from  Equation~\eqref{loss:DA}. ``Trace \& pseudo" uses Equation~\eqref{loss:DA} as the adaptation loss.
Figure~\ref{fig:study_4} shows that ``trace-only" outperforms ``pseudo-only" by a large margin in the first 5 epochs. This superiority contributes to the tight domain gap provided by the frozen many classifiers that 
$G$ must be adapted to make all classifiers in agreement as close as possible. On the other hand, ``pseudo-only" only relies on the target clustering information leveraging from the pretrained source model but neglecting the domain gap. However, pseudo labels do provide global distribution information of the target domain which can be amended in our DAMC framework. Table~\ref{Experiment:study_4} lists the specific accuracy of each category at epoch 15.  Note that for category ``knife", ``pseudo-only" performs much worse than ``trace-only", but the overall performance of ``pseudo-only" is good since it uses the global distribution information of the target domain. Then the best solution is to combine the two losses together as ``trace \& pseudo" which demonstrate the best performance shown in Figure~\ref{fig:study_4} and Table~\ref{Experiment:study_4}. 

\subsubsection{The Effect of Model Selection}

\begin{table*}[!t] 
\caption{Validation of the model selection strategy on category ``train" of VisDA.}
\vskip -0.1in
  \label{Experiment:study_2}
  \begin{center}
  \scalebox{0.65}{
    \begin{tabular}{c|c c c c c c c c c c c c|c}
      \hline
      Train & $C_1$ & $C_2$ & $C_3$ & $C_4$ & $C_5$ & $C_6$ & $C_7$ & $C_8$ & $C_9$ & $C_{10}$ & $C_{11}$ & $C_{12}$ & Acc at Ep.1\\
      \hline
      \rowcolor{lightgray} \cellcolor{white} Model 0 & 99.8 & 99.7 & 99.8 & \cellcolor{white} 100.0 & 99.8 & 99.2 & 99.8 & 99.9 & 99.9 & 99.9 & 99.7 & 99.7 & \cellcolor{white} 82.129\\
      Model 1 & 100.0 & \cellcolor{lightgray} 99.9 & 100.0 & 100.0 & 100.0 & \cellcolor{lightgray} 99.8 & 100.0 & 100.0 & \cellcolor{lightgray} 99.9 & 100.0 & \cellcolor{lightgray} 99.5 & \cellcolor{lightgray} 99.9 & \textbf{84.561}\\
      \rowcolor{lightgray} \cellcolor{white} Model 2 & 99.9 & 99.8 & 99.6 & 99.8 & 99.2 & 99.8 & 99.6 & 99.3 & 99.6 & 99.7 & 99.6 & 99.2 & \cellcolor{white}73.371\\
      \rowcolor{lightgray} \cellcolor{white} Model 3 & 99.8 & 99.7 & \cellcolor{white}100.0 & \cellcolor{white}100.0 & 99.8 & 99.7 & 99.8 & 99.7 & 99.8 & 99.8 & 99.9 & 99.7 & \cellcolor{white}79.969\\
      \underline{Model 4} & 100.0 & 100.0 & 100.0 & 100.0 & 100.0 & \cellcolor{lightgray} 99.9 & 100.0 & 100.0 & 100.0 & 100.0 & \cellcolor{lightgray} 99.8 & 100.0 & \underline{83.121}\\
      \hline
    \end{tabular}
    }
  \end{center}
\end{table*} 

A model selection strategy is proposed in~\ref{sec:model_sel} to select the best model that classifiers are pushed away from each pair whilst ensuring the correctness of the decision boundaries.
We investigate a particular category ``train" in VisDA by comparing five 12-classifier models during the source domain training process.
The accuracy of category ``train" at the 1st adaption epoch is reported in Table~\ref{Experiment:study_2}.
Model 3 performs the worst in this category because all classifiers predict less than 100\% for source training data.
This situation means that all classifiers are placed on the decision boundary of category ``train" but neglecting the correctness.
On the other hand, Model 1 and Model 4 are much better, because there are several perfect classifiers which guarantee the correctness and a few ``imperfect" classifiers which help to maximize the disagreement. 

\section{Conclusion}
\label{sec:conclusion}
This paper explores why and how to use many classifiers for source free UDA. We provide the theoretical analysis from domain gap estimation. Many classifiers do tight the upper bound of domain gap which in turn tight the transferring target error.
For source free UDA, using many classifiers can maximize the possibility of disagreement in the target domain. In theory, the optimal number of classifiers for a task is equal to the number of categories. But in practice, the number of categories can be up to 1000, such as ImageNet, which hinders to use the theoretical best model. How to determine the best trade-off between the number of classifiers and the number of categories is our future work.

 \bibliographystyle{elsarticle-num} 
 \bibliography{damc}

\newpage
\appendix

\section{Theoretical Support}
We rigorously prove the upper bound of the target error by using many classifiers discrepancy.
From a theoretical point of view, for the c-classification problem, using more classifiers can obtain a higher accuracy rate than using the bi-classifier.
Based on the theory of upper bound of error in learning from different domains~\cite{ben2010theory}, we further generalize the theory from $\mathcal{H}\Delta\mathcal{H}$-divergence to $\cap \mathcal{H}\Delta\mathcal{H}$-divergence.

\begin{lemma}[$d_{\mathcal{H}\Delta\mathcal{H}}$-divergence~\cite{ben2010theory}]
Suppose $\mathcal{H} $ is a hypothesis space, and $\mathcal{H}\Delta\mathcal{H}$ is a symmetric difference hypothesis space, $g\in\mathcal{H}\Delta\mathcal{H}, g(x)=h(x)\oplus h^{\prime}(x)$.
When $k=2,c=2$, the $\mathcal{H}\Delta\mathcal{H}$-distance between any two distributions $\mathcal{D}_s$ and $\mathcal{D}_t$ is:
\begin{equation}
    {d_{c=2}^{k=2}}_{\mathcal{H}\Delta\mathcal{H}}(\mathcal{D}_S,\mathcal{D}_T)=2\sup _{g\in \mathcal{H}\Delta\mathcal{H}}|P_{\mathcal{D}_s}[I(g)]-P_{\mathcal{D}_T}[I(g)]|
\end{equation}
For any hypotheses $h,h^\prime\in \mathcal{H}$,
\begin{equation}
    |\epsilon_s(h,h^\prime)-\epsilon_t(h,h^\prime)|\le \frac{1}{2}{d_{c=2}^{k=2}}_{\mathcal{H}\Delta\mathcal{H}}(\mathcal{D}_S,\mathcal{D}_T)
\end{equation}
where $\epsilon_s(h,h^\prime)$ and $\epsilon_t(h,h^\prime)$ is the empirical error of the source and target domain.
\end{lemma}
\begin{table}[H]
  \begin{center}
    \begin{tabular}{| c | c c |}
      \hline
      & 0 & 1\\
      \hline
      0 & 1 & 0\\
      1 & 0 & 1\\
      \hline
    \end{tabular}
    \caption{Disagreement table of bi-classifier for 2-category task.}
      \label{appendix:classifier-2}
  \end{center}
\end{table}
For the bi-classifier, the domain gap between the source and target domain is measured by maximizing the distance between hypotheses.
The disagreement table for the 2-category task using the bi-classifier is as Table~\ref{appendix:classifier-2}.
"0" represents the disagreement of predictions between $h$ and $h^\prime$, and ``1" represents the agreement.
The ratio of disagreement in all results is $P_{c=2}^{k=2}=\dfrac{1}{2}$.
And when $c>2$, as shown in Table~\ref{appendix:DT_classifier_c}, the ratio of disagreement is $P_c^{k=2}=\dfrac{c(c-1)}{c^2}=\dfrac{c-1}{c}$ for c-category task.
\begin{table}[H]
  \begin{center}
    \begin{tabular}{| c | c c c c c|}
      \hline
      & 0 & 1 & 2 & ... & c\\
      \hline
      0 & 1 & 0 & 0 & ... & 0\\
      1 & 0 & 1 & 0 & ... & 0\\
      2 & 0 & 0 & 1 & ... & 0\\
      ... & ... & ... & ... & ... & ...\\
      c & 0 & 0 & 0 & ... & 1\\
      \hline
    \end{tabular}
    \caption{Disagreement table of bi-classifier model for c-category task.}
    \label{appendix:DT_classifier_c}
  \end{center}
\end{table}

As the number of categories increases, the ratio of disagreement between bi-classifier becomes more and more deterministic since $\lim_{c\to\infty}P_{c}^{k=2}=1$.

In view of domain gap estimation by maximizing the discrepancy between bi-classifier~\cite{ben2010theory, saito2018maximum}, the deterministic disagreement of bi-classifier leads to a loose upper bound of domain gap.
This motivates us to explore whether more classifiers help to tighten the upper bound of domain gap estimation.
First, we extend the $\mathcal{H}\Delta\mathcal{H}$ space to $\cap \mathcal{H}\Delta\mathcal{H}$ by combining $\mathcal{H}\Delta\mathcal{H}$ in pairs to accommodate our many classifiers model.
To facilitate calculation, we introduce the soft version of $\cap \mathcal{H}\Delta\mathcal{H}$ and the induced $d_{\cap\mathcal{H} \Delta\mathcal{H}}$-divergence. 
\begin{definition}[Soft $\cap \mathcal{H}\Delta\mathcal{H}$-space]
\label{appendix:multiple_hypothesis_space}
For a hypothesis set $H_k\subset \mathcal{H}$, where $H_k$ is a set of $k$ hypotheses, the multiple difference hypothesis space set $\cap \mathcal{H}\Delta\mathcal{H}$ is the intersection of symmetric difference hypotheses space $\mathcal{H}\Delta\mathcal{H}$.
\begin{equation}
    g\in \cap \mathcal{H}\Delta\mathcal{H}\implies g=\frac{\sum h_i\oplus h_j}{C_k^2},~\forall (h_i,h_j)\in \dbinom{H_k}{2}
\end{equation}
where $C_k^2$ is the number of pairs of classifiers from $\mathcal{H}_k$. $g$ is defined as the expectation of XOR function for pairs of classifiers.
\end{definition}
\begin{definition}[$d_{\cap\mathcal{H} \Delta\mathcal{H}}$-divergence]\label{appendix:d_manyclassifiers}
For $\forall g\in \cap \mathcal{H}\Delta\mathcal{H}$ in the Definition~\ref{appendix:multiple_hypothesis_space} of soft $\cap \mathcal{H}\Delta\mathcal{H}$-space, when there are $k$ hypotheses, the $d_{\cap\mathcal{H} \Delta\mathcal{H}}$-distance is:
\begin{equation}
    \begin{split}
      d_c^{k}(\mathcal{D}_S,\mathcal{D}_T)=2\sup _{g\in \cap \mathcal{H}\Delta\mathcal{H}}|P_{\mathcal{D}_S}[I(g)]-P_{\mathcal{D}_T}[I(g)]|
    \end{split}
\end{equation}
\end{definition}
When using many classifiers, the domain gap is the maximization of the expectation that hypotheses are disagreed across multiple hypotheses. The Definition~\ref{appendix:d_manyclassifiers} of $d_{\cap\mathcal{H} \Delta\mathcal{H}}$-divergence requires access to both source and target domain to touch domain boundaries. But for source free UDA, when pre-training in the source domain, the classifiers need to be pushed away as far as possible, 
 and in the ideal case, some classifiers need to touch the decision boundary whilst ensuring accurate classification. This placement of classifier guarantees that the gap between the source domain and the unknown target domain can be measured more stable by increasing the number of classifiers.

Next, we will demonstrate the superiority of using more classifiers and the optimal number of classifiers by disagreement between multiple hypotheses.
\begin{definition}[Disagreement of hypothesis set $H_k$]
For $\forall h_i,h_j\in H_k$, the disagreement $\epsilon(H_k,H_k)$ is expectation of conjunction result of all hypothetical combinations $(h_i,h_j)\in \binom{H_k}{2}$.
\begin{equation}
    \epsilon(H_k,H_k)=E_{x\sim D} [\frac{\sum h_i\oplus h_j}{C_k^2}] 
\end{equation}
\end{definition}
Suppose that the number of categories is $c$, when $k\in [2,c]$,
\begin{scriptsize}
  \begin{equation}
    \dfrac{P_c^{k}}{P_c^{k-1}}=\dfrac{A_c^k/c^k}{A_c^{k-1}/c^{k-1}}=\dfrac{c-k+1}{c}\in [\dfrac{1}{c},\dfrac{c-1}{c}]\le 1
  \end{equation}
\end{scriptsize}
where $A^k_c$ denotes the number of $k$-permutations of $c$. Then using more classifiers leads to a tighter ratio of disagreement. Note that in case $k > c$, the extra $k-c$ classifiers can not bring more disagreements. For example, if we use 3-classifier model for the 2-classification problem, we will have $P_2^3 = P_2^2 = \frac{1}{2}$ because the three classifiers can be grouped into three groups of bi-classifier in which the ratio of disagreement all equals $\frac{1}{2}$. Obviously,  we have $P_c^{c+1} = P_c^c$.
\begin{proposition}
\label{appendix:law}
For c-category classification task, if $k=c$, $P_c^{k-1}=\dfrac{c}{n-k+1}P_c^k$ and $P_c^{k+1}=P_c^{k}$, that is, (k+1)-classifier model provides the same upper bound estimation as k-classifier model.
\begin{eqnarray}
    {d_c^k}_{\mathcal{H}\Delta\mathcal{H}}(\mathcal{D}_S,\mathcal{D}_T)&=&\dfrac{c-k+1}{c}{d_c^{k-1}}_{\mathcal{H}\Delta\mathcal{H}}(\mathcal{D}_S,\mathcal{D}_T) \quad (if ~k\le c)\\
    {d_c^k}_{\mathcal{H}\Delta\mathcal{H}}(\mathcal{D}_S,\mathcal{D}_T)&=&{d_c^{k+1}}_{\mathcal{H}\Delta\mathcal{H}}(\mathcal{D}_S,\mathcal{D}_T), \quad (if~k\ge c)
\end{eqnarray}
where $\dfrac{c-k+1}{c} \in [\dfrac{1}{c},1)$.
\end{proposition}
\begin{lemma}[Target error bound~\cite{ben2010theory}]
\label{appendix:tgt_error}
Suppose $\mathcal{H}\varDelta \mathcal{H}$ is the hypothesis space. If $U_S, U_T$ are unlabeled samples from $\mathcal{D}_S$ and $\mathcal{D}_T$, whose size are both $m^\prime$, for $\forall \delta \in (0,1), h_i,h_j\in \mathcal{H}$, which the probability is $1-\delta$ at least:
\begin{equation}
    \epsilon_t(h)\le \epsilon_s(h)+\dfrac{1}{2}\hat{d}_{\mathcal{H}\Delta\mathcal{H}}(\mathcal{U}_S,\mathcal{U}_T)+4\sqrt{\dfrac{2d\log (2m^\prime)+\log (\dfrac{2}{\delta})}{m^\prime}}+\lambda
\end{equation}
\end{lemma}
From Lemma~\ref{appendix:tgt_error}, we can derive the target error bound for multiple hypotheses.
\begin{proposition}[Target error bound for $\cap \mathcal{H}\varDelta \mathcal{H}$]
\label{appendix:tat_error_for_many_cls}
Suppose $\cap \mathcal{H}\varDelta \mathcal{H}$ is the hypothesis space. If $U_S, U_T$ are unlabeled samples from $\mathcal{D}_S$ and $\mathcal{D}_T$, whose size are both $m^\prime$, for $\forall \delta \in (0,1), H_k\subset \mathcal{H}$, which the probability is $1-\delta$ at least:
\begin{equation}
    \epsilon_t(H_k)\le \epsilon_s(H_k)+\dfrac{1}{2}\hat{d}^{k}_{\cap \mathcal{H}\Delta\mathcal{H}}(\mathcal{U}_S,\mathcal{U}_T)+4\sqrt{\dfrac{2d\log (2m^\prime)+\log (\dfrac{2}{\delta})}{m^\prime}}+\lambda
\end{equation}
\begin{proof}\renewcommand{\qedsymbol}{}
\begin{equation}
    \begin{split}
      \epsilon_T(H_k)&\le \epsilon_T(h^*)+\epsilon_T(H_k,h^*)\\
      &\le \epsilon_T(h^*)+\epsilon_S(H_k,h^*)+|\epsilon_T(H_k,h^*)-\epsilon_S(H_k,h^*)|\\
      &\le \epsilon_T(h^*)+\epsilon_S(H_k,h^*)+\dfrac{1}{2}d^{(k)}_{\cap\mathcal{H}\Delta\mathcal{H}}\\
      &\le \epsilon_T(h^*)+\epsilon_S(h^*)+\epsilon_S(H_k) +\dfrac{1}{2}d^{(k)}_{\cap\mathcal{H}\Delta\mathcal{H}} \\
      &= \epsilon_S(H_k)+\dfrac{1}{2}d^{(k)}_{\cap\mathcal{H}\Delta\mathcal{H}}+\lambda\\
      &\le \epsilon_S(H_k)+\dfrac{1}{2}\hat{d}^{(k)}_{\cap\mathcal{H}\Delta\mathcal{H}}(\mathcal{U}_S,\mathcal{U}_T)+4\sqrt{\dfrac{2d\log (2m^\prime)+\log (\dfrac{2}{\delta})}{m^\prime}}+\lambda
    \end{split}
\end{equation}
\end{proof}
\end{proposition}
From the Proposition~\ref{appendix:law} and the Proposition~\ref{appendix:tat_error_for_many_cls}, we can draw proposition about the optimal number of classifiers.

\begin{proposition}
\label{appendix:the_optimal_number_of_cls}
For the c-category classification problem, the upper error bound of c-classifier is the tightest, which means that c-classifier is the best choice.
\begin{proof}\renewcommand{\qedsymbol}{}
Suppose $\eta_t(H_k)$ is the upper error bound of $\epsilon_t(H_k)$, that is,
\begin{equation}
    \eta_t(H_k)= \epsilon_s(H_k)+\dfrac{1}{2}\hat{d}^{k}_{\cap \mathcal{H}\Delta\mathcal{H}}(\mathcal{U}_S,\mathcal{U}_T)+4\sqrt{\dfrac{2d\log (2m^\prime)+\log (\dfrac{2}{\delta})}{m^\prime}}+\lambda
\end{equation}
When $2\le k\le c$,
\begin{equation}
      \eta_t(H_k)-\eta_t(H_{k-1})\approx \dfrac{1}{2}\hat{d}^{k}_{\cap \mathcal{H}\Delta\mathcal{H}}(\mathcal{U}_S,\mathcal{U}_T)-\dfrac{1}{2}\hat{d}^{k-1}_{\cap \mathcal{H}\Delta\mathcal{H}}(\mathcal{U}_S,\mathcal{U}_T)< 0
\end{equation}
while $k>c$, $\eta_t(H_k)=\eta_t(H_{k-1})$.
Thus, $\eta_t(H_k)\le \eta_t(H_{k-1})$, which means that using more classifiers own a tighter upper error bound.
\end{proof}
\end{proposition}
Therefore, for the c-category classification problem, the upper error bound of c-classifier is the tightest and the c-classifier is the best choice.
%
In practice, if the number of categories $c$ is large, using so many classifiers bound the upper bound tightest though, it would cost too much computation. There is a trade-off between the accuracy of domain gap estimation and computation.
For practical source free UDA tasks, if $c$ is large, we propose to use $k$ classifiers which induce the ratio of disagreement  \begin{scriptsize}$P_c^{k}=A_c^k/c^k\approx \epsilon$\end{scriptsize}. Here $\epsilon$ is a predefined threshold of disagreement ratio which controls how tight of the domain gap estimation our model should achieve.

\section{Network Structure Settings and Hyper-parameters}
\subsection{Network Structure Settings}
In this work, for Digits tasks, we adopt the CNN architecture used in \cite{ganin2015unsupervised} and \cite{bousmalis2017unsupervised} for feature generation, and use fully connected layers as classifiers.
For medium-scale and large-scale benchmarks, we adopt ResNet~\cite{he2016deep} as pre-trained base models for VisDA (ResNet-101) and Office-Home tasks (ResNet-50).
Except the last fully connected layer, ResNet is used as a feature generator. Bottleneck layer is used to squeeze the feature dimension to 256. For classifiers, we use three fully connected layers for VisDA and two fully connected layers for Office-Home.
Batch normalization~\cite{pmlr-v37-ioffe15} layers are inserted into the classifiers for each task.

Table~\ref{appendix:Network Structure Settings} shows the specific network structure settings for Digits, VisDA and Office-Home.
\begin{table}[H]\scriptsize
  \begin{center}
    \begin{tabular}{| c | c | c c |}
      \hline
      Tasks & Source & Generator & Classifiers\\
      \hline
      Digits&USPS/MNIST &CNN from scratch& 2 fully connected\\
      Digits&SVHN &CNN from scratch&3 fully connected\\
      VisDA&- &ResNet-101 with bottleneck layer&3 fully connected layers\\
      Office-Home&- &ResNet-50 with bottleneck layer&2 fully connected layers\\
      \hline
    \end{tabular}
    \caption{Network Structure Settings for Digits, VisDA and Office-Home}
    \label{appendix:Network Structure Settings}
  \end{center}
\end{table}
\subsection{Training Details}
Our models are optimized by SGD with momentum\cite{sutskever2013importance} and the learning rate is scheduled by the same scheduler as~\cite{ganin2015unsupervised}.
Weight decay is also applied with ${5.0}\times{10}^{-4}$ during training.
\begin{scriptsize}
\begin{equation}
    \eta = \eta_0 \cdot \left(1 + \frac{10\cdot iter}{maxIter}\right)^{-0.75}
\end{equation}
\end{scriptsize}

Table~\ref{appendix:learning-rate} shows the specific learning rate $\eta_0$ settings for Digits, VisDA and Office-Home.
\begin{table}[H] \scriptsize
  \begin{center} 
    \begin{tabular}{| c|c |c |c|c|}
      \hline
       $\eta_0$&Digits& VisDA & Office-Home& Note\\
      \hline
      C (Pretrain) &2e-2&1e-2 &1e-2 &Initial learning rate of C in pre-training.\\
      G (Pretrain)&2e-2&1e-4 &1e-4&Initial learning rate of G in pre-training.\\
      Botttleneck (Pretrain)&-& 1e-2 &1e-2&Initial learning rate of bottleneck in pre-training.\\
      \hline
      C (DA)&2e-2&1e-3& 1e-3&Initial learning rate of C in Domain Adaptation.\\
      G (DA) &2e-2&1e-3 &1e-3&Initial learning rate of G in Domain Adaptation.\\
      Bottleneck (DA) &-&1e-3&1e-3&Initial learning rate of bottleneck in Domain Adaptation.\\
      \hline
    \end{tabular}
    \caption{Learning Rate $\eta_0$ for Tasks.}
    \label{appendix:learning-rate}
  \end{center}
\end{table}

\subsection{Hyper-parameters}
Table~\ref{appendix:Hyper-parameters} shows the hyper-parameters used in our experiments. $\alpha_s$ is the coefficient of the adversarial loss for pushing classifiers away in pre-training. $\tau$ is the prescribed threshold that should be guaranteed by the smallest discrepancy among all pairs of classifiers in pre-training. 
$\alpha_t$ is the coefficient for pair of trace loss in DA. $\beta$ is the coefficient of pseudo labeling loss in DA. $p_i$ indicates the number of epochs to re-calculate the pseudo labels. $p_s$ indicates at which epoch the pseudo labeling loss begins to be used.
\begin{table}[H]\scriptsize
  \begin{center}
    \begin{tabular}{| c |c |c|c|}
      \hline
       Hyper-parameters& VisDA & Office-Home& Note\\
      \hline
      $\alpha_s$ &0.3 &1 &The coefficient of the adversarial loss in pre-training.\\
      $\tau$ &0.4 &0.05 & Threshold of the worst optimization in pre-training.\\\
      $\alpha_t$ &0.1& 0.5 &The coefficient of the pair of trace loss in DA.\\
      $\beta$ & 0.01 &0.1&The coefficient of pseudo labeling loss.\\
      $p_i$ & 2 &2& Pseudo labels are re-calculated at every $p_i$ epochs.\\
      $p_s$ &2 &1& When the pseudo labeling loss starts to be used.\\
      \hline
    \end{tabular}
    \caption{Hyper-parameters for VisDA and Office-Home.}
    \label{appendix:Hyper-parameters}
  \end{center}
\end{table}






\end{document}